%% file: main.tex
\pdfoutput=1
\documentclass[10pt,twocolumn,letterpaper]{article}
\usepackage[pagenumbers]{cvpr}
\input{preamble}
\definecolor{cvprblue}{rgb}{0.21,0.49,0.74}
\usepackage[pagebackref,breaklinks,colorlinks,allcolors=cvprblue]{hyperref}
\def\confName{CVPR}
\def\confYear{2027}
\title{RefreshGrad: An Online Adaptive Data-sampling Method for Diffusion-based Visuomotor Learning}
\author{
Dijia Zhan$^{1,*}$ \quad
Jinyi Li$^{1,*}$ \quad
Xuemiao Xu$^{1,\dagger}$ \quad
Yan Zhou$^{1}$ \quad
Yu Liu$^{2}$ \quad
Shaoshan Liu$^{3}$ \quad
Jie Tang$^{1,\dagger}$\\[4pt]
$^{1}$South China University of Technology \quad
$^{2}$Tsinghua University \\
$^{3}$The Chinese University of Hong Kong, Shenzhen\\[2pt]
{\small $^{*}$Equal contribution \quad $^{\dagger}$Corresponding authors}
}
\begin{document}

\twocolumn[{
    \renewcommand\twocolumn[1][]{#1}
    \maketitle
    \begin{center}
        \captionsetup{type=figure}
        \includegraphics[width=0.95\textwidth]{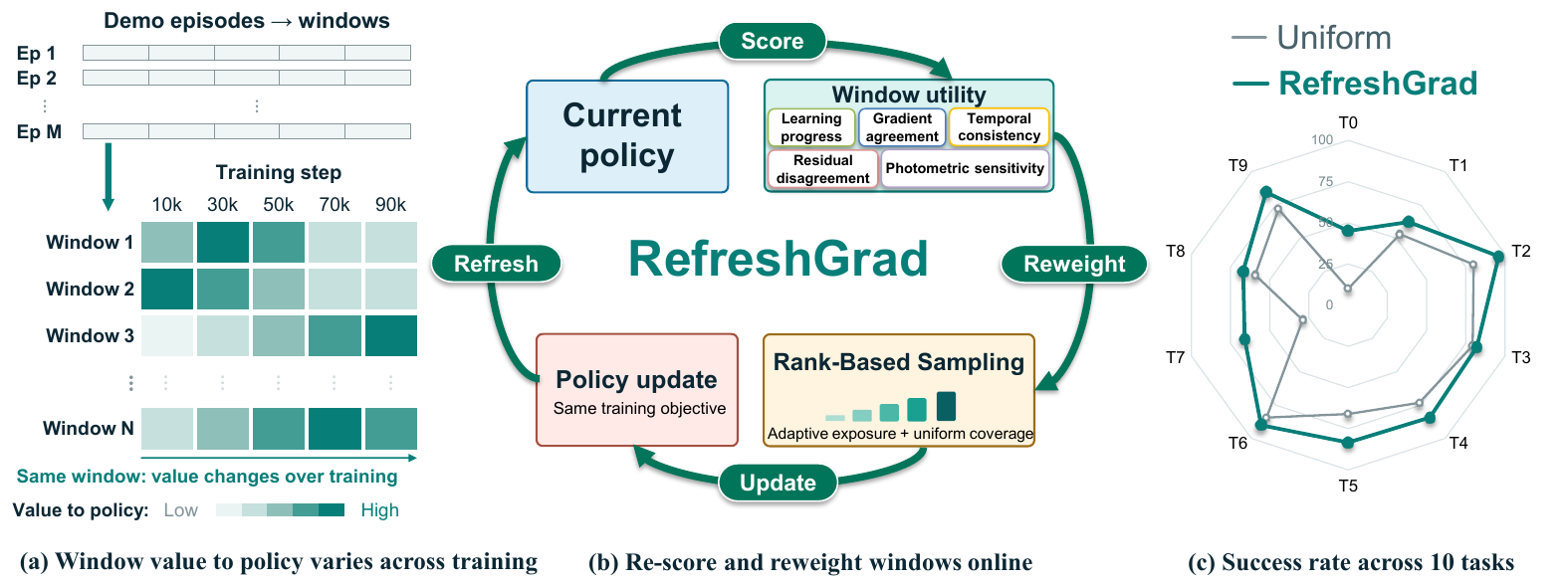}
        \captionof{figure}{\textbf{Unequal and dynamic window value to the policy motivates online re-scoring and reweighting.} \textbf{(a)} Windows sliced from demonstration episodes have unequal value to the current policy at the same training step; each window's value changes over training (illustrative). \textbf{(b)} \method periodically re-scores windows with the current policy using policy-native signals, applies rank-based sampling with uniform coverage, and updates the policy under the same training objective. \textbf{(c)} \method improves perturbed success over Uniform on all ten LIBERO-Plus tasks, raising the task average from 60.7\% to 76.1\% (+15.4 pp).}
        \label{fig:refreshgrad-teaser}
    \end{center}
}]

\input{sec/0_abstract}
\input{sec/1_intro}
\input{sec/2_related}
\input{sec/3_observations}
\input{sec/3_method}
\input{sec/4_experiments}

\section{Conclusion}
We presented \method, an online adaptive data-sampling framework for demonstration-based policy learning driven by the non-uniform and evolving value of temporal windows during training. By integrating window-level learning progress with cross-episode gradient agreement, temporal prediction consistency, residual disagreement, and photometric sensitivity through rank-based sampling and a uniform coverage floor, \method emphasizes learnable transitions while suppressing redundant or noisy segments. Across a ten-task LIBERO-Plus panel, \method boosts average rollout success from 60.7\% (Uniform) and 68.2\% (FAKTUAL) to \textbf{76.1\%} under visual perturbations and from 57.4\% (Uniform) and 73.3\% (FAKTUAL) to \textbf{81.3\%} in clean scenes, with a 19\% median training scoring overhead and zero inference cost.

\paragraph{Limitations and future work.} To cleanly isolate training-time window allocation from language-grounding confounds under a controlled budget, our empirical study focuses on an action-chunking Diffusion Policy backbone evaluated across the stratified LIBERO-Plus panel (4,000 rollouts per method). Extending \method to large-scale language-conditioned VLA backbones and physical robot platforms are promising directions for future work.

\section*{Acknowledgments}
This work was supported by Guangdong Provincial Natural Science Foundation for Outstanding Youth Team Project (No.~2024B1515040010), National Natural Science Foundation of China under Grant U23A20391, 62372188, Guangdong Natural Science Foundation under Grant 2024A1515010100, and Pivotal Technology Research Project of the Hengqin Guangdong-Macao In-Depth Cooperation Zone under Grant 2530004025492.

{\small
\bibliographystyle{ieeenat_fullname}
\bibliography{main}
}
\input{sec/appendix}
\end{document}

%% file: preamble.tex
\usepackage[T1]{fontenc}

\usepackage{booktabs}
\usepackage{colortbl}
\definecolor{rgfull}{HTML}{F4F4F4}
\definecolor{rgfaktual}{HTML}{F0F4F9}
\definecolor{rgmethod}{HTML}{FCEEEE}
\usepackage{multirow}
\usepackage{microtype}
\usepackage{float}
\usepackage{mathtools}
\usepackage{enumitem}
\usepackage{tikz}
\usetikzlibrary{arrows.meta,positioning,patterns}
\setlist{nosep,leftmargin=*}
\newcommand{\method}{RefreshGrad\xspace}

\tikzset{
  flowbox/.style={draw, rounded corners, align=center, minimum width=3.0cm,
    minimum height=0.62cm, font=\footnotesize, inner sep=3pt},
  flowarrow/.style={-{Latex[length=1.5mm]}, semithick}
}

%% file: sec/0_abstract.tex
\begin{abstract}
Training vision--language--action (VLA) policies from demonstrations typically samples trajectory segments uniformly or relies on static offline data curation before training. However, we find that demonstration data vary sharply in supervision value across temporal windows, and that each window's value to the policy evolves substantially during training rather than being fixed or captured by raw loss alone. Motivated by this finding, we propose \method, an online adaptive data-sampling framework that periodically scores temporal windows from the evolving policy using five policy-native signals and dynamically adjusts their sampling weights during training. \method focuses optimization on informative, learnable windows without altering the policy architecture, training objective, or inference cost. Across ten LIBERO-Plus manipulation tasks, \method boosts average success from 60.7\% (Uniform) and 68.2\% (FAKTUAL) to \textbf{76.1\%} under visual perturbations, and from 57.4\% and 73.3\% to \textbf{81.3\%} in clean scenes.
\end{abstract}

%% file: sec/1_intro.tex
\section{Introduction}

Adapting visuomotor and vision--language--action (VLA) policies to downstream manipulation tasks relies heavily on demonstration supervision \cite{brohan2023rt2,kim2024openvla,chi2023diffusion}. However, a fundamental yet under-appreciated property of demonstration datasets is the severe \emph{disparity in data value} across temporal windows. Across episodes, demonstrations inevitably mix clean executions with visual corruptions---such as dim lighting, sensor blur, and background texture shifts---as well as demonstrator inconsistency. Even within a single successful episode, temporal segments vary sharply in supervision utility: repetitive free-space reaching motions offer diminishing returns once mastered, fine contact and articulation transitions demand concentrated optimization, and hesitant or occluded intervals can induce conflicting parameter updates. Treating all segments equally under uniform sampling wastes optimization budget on redundant motions while allowing corrupted or inconsistent windows to degrade policy learning.

Our analysis shows that training value is both window-dependent and policy-dependent: windows within the same episode contribute unequally (\cref{fig:refreshgrad-teaser}(a)), and per-window signals that are stable between adjacent refreshes decorrelate steadily as training proceeds, under both RefreshGrad and uniform sampling (\cref{sec:appendix-value-dynamics}). A fixed assessment of data quality therefore cannot fully capture a window's current learning value. A clean but already-mastered window may offer limited additional supervision, whereas an imperfect window may still contain learnable transitions. This observation motivates adapting exposure to the evolving learner, beyond static offline selection \cite{sirigiri2026faktual,dass2025datamil}. Since high loss can also reflect corruption or inconsistent actions, this adaptation requires feedback about both learning progress and prediction consistency.

Motivated by this insight, we propose \method, an online adaptive data-sampling framework that repeatedly estimates window utility from the current policy and reallocates training exposure accordingly (\cref{fig:refreshgrad-teaser}(b)). Following a uniform warmup, \method scores windows using learning progress, cross-episode gradient agreement, temporal prediction consistency, residual disagreement, and photometric sensitivity under deterministic diffusion probes. These complementary signals distinguish actively learnable windows from redundant or unreliable supervision. Robust normalization and temporal smoothing yield comparable, stable utilities, which a rank-based sampler combines with a uniform coverage floor to update anchor sampling probabilities. Alternating scoring with policy updates allows exposure to track changing window value while preserving nonzero support for every valid segment. The procedure requires no external scorer or rollouts and leaves the policy architecture, per-sample loss, and inference pipeline unchanged.

We instantiate \method on an action-chunking Diffusion Policy backbone \cite{chi2023diffusion} and evaluate it across ten LIBERO-Plus tasks. As summarized in \cref{fig:refreshgrad-teaser}(c), \method improves perturbed success over Uniform on all ten tasks, raising the average from 60.7\% to 76.1\%. It also exceeds the 68.2\% achieved by FAKTUAL \cite{sirigiri2026faktual} and improves clean success from 57.4\% to 81.3\%, supporting the effectiveness of online adaptive window allocation.

Our contributions are threefold:
\begin{itemize}
\item \textbf{Characterization of demonstration data value.} We reveal that temporal-window value is non-uniform and evolves with the learner, and identify its principal sources as learning progress, cross-episode gradient agreement, temporal prediction consistency, residual disagreement, and photometric sensitivity. This provides a policy-native account of why different windows contribute unequally.
\item \textbf{Data-value-driven online window allocation.} We introduce \method to periodically estimate these value cues from the current policy and adapt training exposure through rank-based sampling with a uniform coverage floor, without environment rollouts or inference overhead.
\item \textbf{Empirical and mechanistic validation.} We evaluate \method across ten LIBERO-Plus tasks and show gains in both perturbed and clean success. Training-trace diagnostics, sampler controls, and signal ablations support multi-cue valuation and periodic online adaptation.
\end{itemize}

%% file: sec/2_related.tex
\section{Related Work}

\paragraph{Visuomotor and vision--language--action policies.} Modern robot manipulation policies map visual observations---often paired with language instructions or multimodal prompts---directly to control actions \cite{brohan2023rt1,brohan2023rt2,kim2024openvla,jiang2023vima}. At the action-generation level, action-chunking transformers and diffusion policies model temporally extended control sequences \cite{zhao2023act,chi2023diffusion}, building on denoising and score-based generative modeling \cite{ho2020ddpm,song2019score,song2021ddim,nichol2021improved} as well as diffusion formulations for planning and offline reinforcement learning \cite{janner2022diffuser,wang2023diffusionql}. Standardized manipulation benchmarks further emphasize long-horizon temporal structure, task diversity, and robustness to environmental variations \cite{liu2023libero,mees2022calvin,james2020rlbench}. While these advances establish expressive policy architectures and evaluation protocols, they typically rely on uniform sampling over collected demonstrations during training. \method addresses the complementary problem of dynamically allocating training exposure across demonstration windows as optimization progresses, and is instantiated here on an action-chunking Diffusion Policy backbone \cite{chi2023diffusion} to isolate sampling effects from language-grounding confounds.

\paragraph{Data selection and valuation for imitation learning.} Offline imitation learning studies consistently show that demonstration quality, diversity, and composition critically govern downstream policy performance \cite{mandlekar2021robomimic}. In general machine learning, active learning and coreset selection allocate supervision toward informative or representative subsets \cite{cohn1994active,sener2018active}, while gradient approximation, gradient matching, validation-guided selection, and influence functions estimate sample utility from model gradients \cite{mirzasoleiman2020craig,killamsetty2021gradmatch,killamsetty2021glister,koh2017influence}. Extending data curation to robot manipulation, DataMIL leverages datamodel attribution and surrogate losses to select task-relevant demonstrations offline \cite{dass2025datamil}, whereas FAKTUAL introduces a fast, model-free trajectory-kernel entropy metric to curate diverse demonstration subsets without policy access or rollouts \cite{sirigiri2026faktual}. Despite their effectiveness, offline curation and static reweighting schemes fix data importance prior to or early in training: model-free diversity curation does not account for within-trajectory quality disparities or visual corruptions, and static selection cannot adapt as the policy masters simple motions and shifts its learning bottleneck (window-utility rank correlation falls from 0.80 between adjacent refreshes to 0.10 between the first and last refresh). Moreover, trajectory-level hard filtering permanently discards excluded episodes that may still contain valuable contact transitions. In contrast, \method operates at the fine-grained temporal-window level, periodically recomputing priorities online from the evolving policy while preserving nonzero support across all demonstrations.

\paragraph{Online adaptive sampling and robustness cues.} Dynamically adjusting sample priorities during training has a long history, notably in prioritized experience replay (PER), which weights reinforcement learning transitions by temporal-difference error \cite{schaul2016prioritized}. However, in supervised demonstration learning under uneven data quality, prioritizing samples solely by loss magnitude is brittle: our training-trace analysis shows that high window loss correlates negatively with cross-episode gradient consensus and temporal action consistency, reflecting demonstrator noise or visual corruptions rather than learnable skill deficiencies. While overlapping action-chunk averaging has been used at inference time to smooth execution \cite{zhao2023act}, and visual robustness research applies domain randomization or augmentations during training \cite{tobin2017domain,hendrycks2020augmix,hendrycks2019corruptions}, prior work has not unified these properties into an online data-valuation mechanism. Rather than altering training images or adding auxiliary losses, \method is the first to formalize cross-episode gradient agreement, shift-aligned temporal consistency, inter-noise residual disagreement, and read-only photometric sensitivity under deterministic diffusion probes as online window-utility dimensions, separating genuinely learnable transitions from noisy or redundant segments while preserving the native training objective and inference pipeline.

%% file: sec/3_observations.tex
\section{Observations on Window-Level Data Value}
\label{sec:observations}

Building on the window-value disparity in \cref{fig:refreshgrad-teaser}(a), we first analyze theoretically what makes a demonstration window valuable. In diffusion-based imitation learning, raw training loss $\mathcal{L}(\theta;i)$ only reflects instantaneous fitting error, which conflates actively learnable transitions with unlearnable noise. From an optimization perspective, we hypothesize that a window's true training value is governed by five complementary factors spanning three requirements---\emph{active error reduction}, \emph{spatiotemporal supervisory coherence}, and \emph{prediction stability under input perturbations}:

\begin{figure}[!t]
\centering
\includegraphics[width=0.85\linewidth]{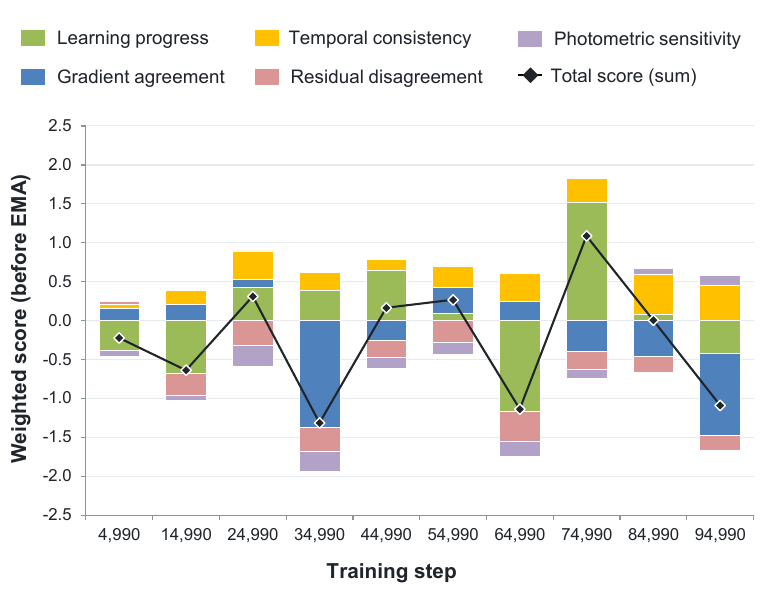}
\caption{\textbf{Empirical verification of the five hypothesized value factors across training refreshes.} Stacked bars show each factor's weighted contribution to a window's pre-EMA utility (black curve). Confirming our theoretical analysis, the five factors neither collapse into a single trend nor remain static: positive learning progress can be counterbalanced by negative gradient agreement when updates conflict across episodes (e.g., step 34,990), while the same window shifts from negative utility (step 64,990) to peak value (step 74,990).}
\label{fig:empirical-value}
\end{figure}

\textbf{(1) Learning progress.} A static loss cannot separate a window where the policy is actively improving from a mastered reaching segment or a persistently noisy transition; only loss reduction across checkpoints indicates where supervision is actively being absorbed. \textbf{(2) Cross-episode gradient agreement.} Even when loss decreases locally, an isolated demonstration window $w$ can mislead the policy if the demonstrator hesitated or took an idiosyncratic path. Since a parameter step $-\eta g_w$ changes the loss of another visually similar window $v$ to first order by $-\eta\langle g_v,g_w\rangle$, positive gradient alignment across episodes is theoretically required for updates to generalize rather than conflict. \textbf{(3) Temporal prediction consistency.} Action-chunking policies predict overlapping future action sequences at adjacent frames; coherent supervision requires adjacent predictions to agree on their shared future horizon rather than inducing jerky control transitions. \textbf{(4) Residual disagreement.} Because diffusion policies denoise stochastic Gaussian inputs, a window where distinct noise realizations produce diverging action reconstructions (relative to current loss scale) indicates high epistemic instability that requires further training exposure. \textbf{(5) Photometric sensitivity.} Finally, task-preserving camera appearance shifts should leave intended actions invariant; high action sensitivity to brightness or contrast shifts reveals reliance on fragile visual shortcuts.

\paragraph{Empirical verification via \cref{fig:empirical-value}.} To test this theoretical hypothesis, we probe demonstration windows across training refreshes using fixed diffusion timesteps and noise banks. \Cref{fig:empirical-value} tracks the five weighted factors and their sum for a representative window over training, directly confirming our hypothesis in three ways. First, the five factors do not collapse into a single dimension: at step 34,990, positive learning progress and temporal consistency are overridden by strongly negative gradient agreement, suppressing a window whose gradient conflicts with broader demonstrations. Second, the dominant factor governing the same window shifts over time, turning a low-utility window at step 64,990 into a high-value window at step 74,990 once gradient conflict subsides and learning progress peaks. Third, dataset-wide refresh statistics corroborate this complementarity: window loss correlates negatively with gradient agreement ($r=-0.31$) and temporal consistency ($r=-0.40$), while loss-normalized residual disagreement ($r=0.13$) and photometric sensitivity ($r=-0.20$; raw shift $r=0.08$) capture nearly orthogonal fragility axes. Consequently, overall window-utility rank correlation drops from $0.80$ between adjacent refreshes to $0.10$ from first to last---and uniform-training logs exhibit the same rank drift without adaptive sampling (\cref{fig:signal-decay}(b))---confirming that window value is both multi-dimensional and time-varying.

%% file: sec/3_method.tex
\begin{figure*}[t]
\centering
\includegraphics[width=1\linewidth]{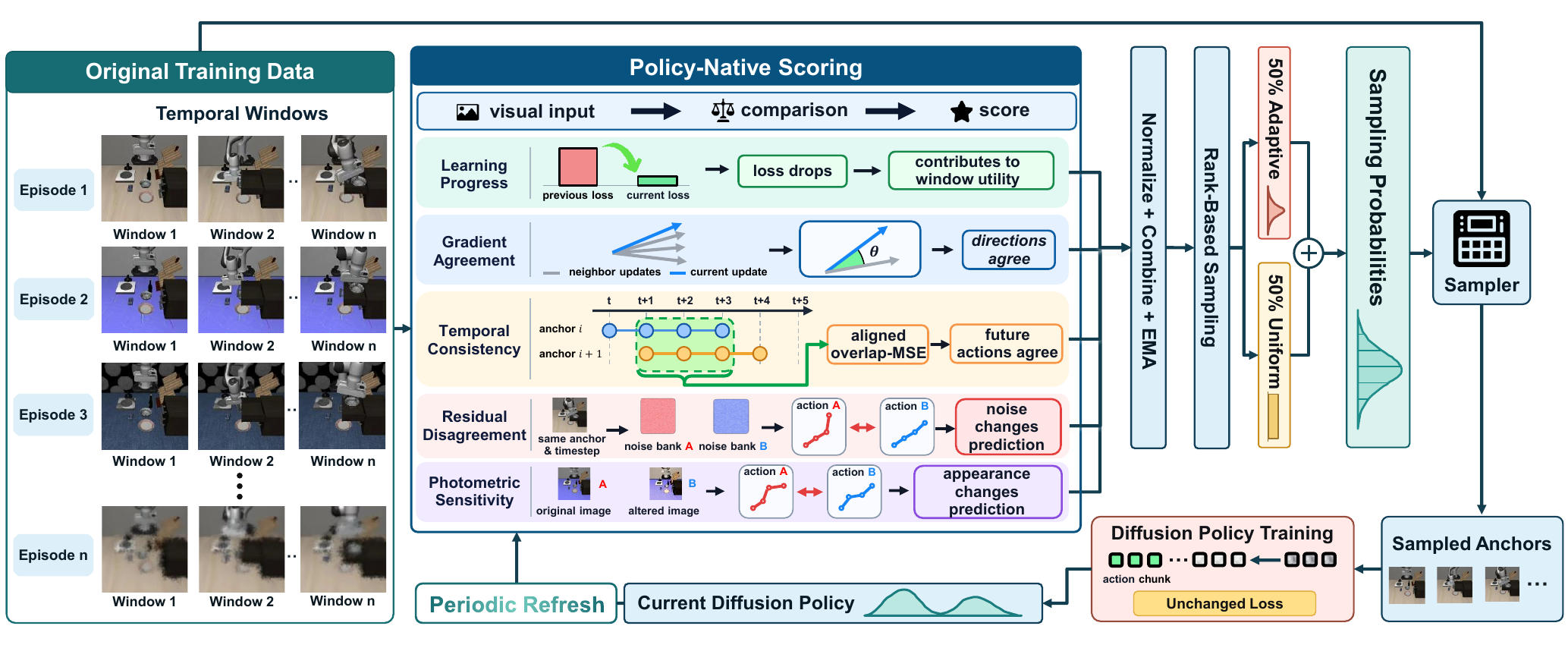}
\caption{\textbf{Overview of the \method alternating training loop.} Periodically during optimization, the evolving policy $\pi_\theta$ evaluates temporal demonstration windows using five policy-native signals: learning progress ($s_j^{\text{prog}}$), gradient agreement ($s_j^{\text{grad}}$), temporal consistency ($s_j^{\text{temp}}$), residual disagreement ($s_j^{\text{res}}$), and photometric sensitivity ($s_j^{\text{photo}}$). Signals are robustly standardized across windows, smoothed via an exponential moving average (EMA), and mapped to anchor sampling probabilities through a mixture of rank-based sampling and a uniform coverage floor, leaving the per-sample diffusion loss and inference pipeline unchanged.}
\label{fig:refreshgrad-pipeline}
\end{figure*}

\section{Method}
\label{sec:method}

\subsection{Problem Setup and Overview}
Motivated by \cref{sec:observations}, we seek to allocate training exposure according to the evolving value of demonstration windows. Each valid anchor $i$ retains an observation history, action chunk, and padding mask. We partition episodes into disjoint windows $W_j=[a_j,b_j)$ of up to $L_w$ consecutive anchors. The problem is to adapt anchor sampling probabilities $p_i$ as the policy $\pi_\theta$ evolves, emphasizing informative windows while preserving coverage and the per-sample objective $\mathcal{L}(\theta;i)$.

We instantiate \method on an action-chunking Diffusion Policy \cite{chi2023diffusion}, minimizing the masked denoising score-matching loss:
\begin{equation}
\mathcal{L}(\theta;i)=\frac{1}{|\mathcal{M}_i|}\sum_{k\in\mathcal{M}_i}\bigl\|\epsilon_\theta(x_i^{m},m)_k-\epsilon_{i,k}\bigr\|_2^2,
\end{equation}
where $m$ is the diffusion timestep, $x_i^{m}$ is the noised action chunk, $k$ indexes future action steps within the chunk, and $\mathcal{M}_i$ is the set of valid (unpadded) steps.

\paragraph{Framework overview.} As shown in \cref{fig:refreshgrad-pipeline}, \method converts the five factors into policy-native window scores, combines them into a smoothed utility, and maps window ranks to anchor sampling probabilities. Following uniform warmup, scoring and reweighting recur every $I_{\text{ref}}$ optimization steps. Policy updates retain the standard denoising loss, while a uniform floor preserves support for all valid anchors.

\subsection{Policy-Native Multi-Cue Utility}
We now instantiate the five factors in \cref{sec:observations} as policy-native scores under fixed, reproducible scoring probes.

\paragraph{Deterministic scoring probes.} To measure these value dimensions without stochastic diffusion-sampling variance, we evaluate each valid anchor $i\in W_j$ at refresh $r$ under a fixed set of deterministic scoring conditions $c=(m,b)\in\mathcal{C}$, formed by pairing representative diffusion timesteps $m\in\mathcal{T}_s$ with $B$ fixed, frame-aligned noise banks $b\in\{1,\dots,B\}$. Holding noise realizations constant across training ensures that changes in evaluated losses and action reconstructions across refreshes reflect genuine evolution of policy parameters $\theta$ rather than resampling noise. Let $\ell_i^{(r)}=|\mathcal{C}|^{-1}\sum_{c\in\mathcal{C}}\mathcal{L}_c(\theta;i)$ denote the mean masked denoising loss of anchor $i$ across $\mathcal{C}$, and $\ell_j^{(r)}=|W_j|^{-1}\sum_{i\in W_j}\ell_i^{(r)}$ the window-level loss. For any $c=(m,b)$, let $\hat{\mathbf a}_{i,c}(k)$ (or $\hat{\mathbf a}_{i,m,b}(k)$) denote the reconstructed action at future timestamp $k$, scaled by $s_m=\sqrt{\bar\alpha_m/(1-\bar\alpha_m)}$ with cumulative noise coefficient $\bar\alpha_m$.

For each window $W_j$, the five scores are indexed by $\mathcal{S}=\{\text{prog},\text{grad},\text{temp},\text{res},\text{photo}\}$. Subscript $i$ denotes an anchor-level quantity and $j$ its window aggregation. Together, they capture active learning, cross-episode and temporal coherence, and stochastic or visual sensitivity.

\paragraph{1. Learning progress ($s_j^{\text{prog}}$).} To identify windows where the policy is actively improving rather than stagnating on unlearnable noise or already-mastered motions, we measure the window-loss reduction since the preceding refresh:
\begin{equation}
s_j^{\text{prog}}=\ell_j^{(r-1)}-\ell_j^{(r)},
\end{equation}
where positive values indicate active error reduction on $W_j$, and $\ell_j^{(0)}$ is recorded during uniform warmup.

\paragraph{2. Cross-episode gradient agreement ($s_j^{\text{grad}}$).} High loss reduction or large gradients on an isolated window can still arise from idiosyncratic demonstrator motion. To verify that a window induces an update supported by broader task demonstrations, let $\mathbf h_j$ denote the mean anchor gradient over $i\in W_j$ (averaged across $c\in\mathcal{C}$ with respect to the action projection head parameters for efficiency). Across the task's demonstration windows, we retrieve up to $M$ windows $\mathcal N_j$ from other episodes with the most similar mean visual-conditioning features. Normalizing each neighbor gradient to unit length and applying a coordinate-wise $\rho$-trimmed mean yields a robust reference direction $\bar{\mathbf h}_j$, from which we compute the cosine similarity:
\begin{equation}
s_j^{\text{grad}}=\frac{\mathbf h_j^\top\bar{\mathbf h}_j}{\|\mathbf h_j\|_2\|\bar{\mathbf h}_j\|_2}.
\end{equation}
Higher $s_j^{\text{grad}}$ indicates that optimizing on $W_j$ aligns updates with visually related demonstrations.

\paragraph{3. Temporal prediction consistency ($s_j^{\text{temp}}$).} Action-chunking policies predict overlapping future control sequences at consecutive timesteps. On coherent segments, adjacent anchors $i$ and $i+1$ within an episode should produce consistent action reconstructions on their overlapping future timestamps $\Omega_i$. Using scaled reconstructions $\hat{\mathbf a}_{i,c}(k)$ across $c\in\mathcal{C}$, we define:
\begin{equation}
\begin{gathered}
s_i^{\text{temp}}=-\frac{1}{|\mathcal{C}||\Omega_i|d_a}\!\sum_{c\in\mathcal{C},\,k\in\Omega_i}\!
\bigl\|\hat{\mathbf a}_{i,c}(k)-\hat{\mathbf a}_{i+1,c}(k)\bigr\|_2^2,\\
s_j^{\text{temp}}=\frac{1}{|\mathcal{P}_j|}\sum_{i\in\mathcal{P}_j}s_i^{\text{temp}},
\end{gathered}
\end{equation}
where $d_a$ is the action dimension and $\mathcal{P}_j$ is the set of valid adjacent pairs $(i,i+1)$ whose later anchor lies in $W_j$ ($\mathcal{P}_j=\emptyset\Rightarrow s_j^{\text{temp}}=0$). Higher $s_j^{\text{temp}}$ favors windows with temporally self-consistent predictions.

\paragraph{4. Residual prediction disagreement ($s_j^{\text{res}}$).} To gauge epistemic prediction instability without being confounded by raw loss magnitude, we compare action reconstructions $\hat{\mathbf a}_{i,m,1}$ and $\hat{\mathbf a}_{i,m,2}$ produced under distinct noise banks at each scoring timestep $m\in\mathcal{T}_s$:
\begin{equation}
\begin{gathered}
s_i^{\text{res}}=\frac{1}{|\mathcal{T}_s|d_a|\mathcal M_i|}\!\sum_{m\in\mathcal{T}_s,\,k\in\mathcal M_i}\!
\bigl\|\hat{\mathbf a}_{i,m,1}(k)-\hat{\mathbf a}_{i,m,2}(k)\bigr\|_2^2,\\
\bar{s}_j^{\text{res}}=\frac{1}{|W_j|}\sum_{i\in W_j}s_i^{\text{res}},\qquad
s_j^{\text{res}}=\log\frac{\bar{s}_j^{\text{res}}+\epsilon}{\ell_j^{(r)}+\epsilon},
\end{gathered}
\end{equation}
where $\epsilon>0$ ensures numerical stability. Normalizing mean inter-bank disagreement $\bar{s}_j^{\text{res}}$ by current window denoising loss $\ell_j^{(r)}$ captures relative reconstruction sensitivity across noise realizations rather than rewarding persistently high-loss noisy anchors.

\begin{table*}[!t]
\centering
\begingroup
\setlength{\fboxsep}{1pt}
\caption{Rollout success rates (\%) across ten LIBERO-Plus tasks under lighting, noise, and texture perturbations and in unperturbed (Clean) scenes. \protect\colorbox{rgfull}{U}: Uniform, \protect\colorbox{rgfaktual}{FK}: FAKTUAL \cite{sirigiri2026faktual}, and \protect\colorbox{rgmethod}{\textbf{RG}}: \method. Pert.\ mean averages the three perturbation families. Best results for each task and condition are bolded.}
\label{tab:refreshgrad-selected-10}
\setlength{\tabcolsep}{0pt}
\renewcommand{\arraystretch}{1.15}
\edef\rgInstructionWidth{\the\dimexpr\linewidth-323pt\relax}
\newcommand{\rgTableCell}[2]{\parbox[c][12.5pt][c]{#1}{\centering #2}}
\footnotesize
\begin{tabular}{@{}>{\raggedright\arraybackslash}m{\rgInstructionWidth} >{\columncolor{rgfull}\centering\arraybackslash}m{21pt} >{\columncolor{rgfaktual}\centering\arraybackslash}m{21pt} >{\columncolor{rgmethod}\centering\arraybackslash}m{21pt} @{\hspace{2pt}} >{\columncolor{rgfull}\centering\arraybackslash}m{21pt} >{\columncolor{rgfaktual}\centering\arraybackslash}m{21pt} >{\columncolor{rgmethod}\centering\arraybackslash}m{21pt} @{\hspace{2pt}} >{\columncolor{rgfull}\centering\arraybackslash}m{21pt} >{\columncolor{rgfaktual}\centering\arraybackslash}m{21pt} >{\columncolor{rgmethod}\centering\arraybackslash}m{21pt} @{\hspace{2pt}} >{\columncolor{rgfull}\centering\arraybackslash}m{21pt} >{\columncolor{rgfaktual}\centering\arraybackslash}m{21pt} >{\columncolor{rgmethod}\centering\arraybackslash}m{21pt} @{\hspace{2pt}} >{\columncolor{rgfull}\centering\arraybackslash}m{21pt} >{\columncolor{rgfaktual}\centering\arraybackslash}m{21pt} >{\columncolor{rgmethod}\centering\arraybackslash}m{21pt}@{}}
\toprule
\multirow{2}{*}{Instruction} & \multicolumn{3}{c}{Lighting (\%)} & \multicolumn{3}{c}{Noise (\%)} & \multicolumn{3}{c}{Texture (\%)} & \multicolumn{3}{c}{Pert.\ mean (\%)} & \multicolumn{3}{c}{Clean (\%)} \\
\cmidrule(lr){2-4}\cmidrule(lr){5-7}\cmidrule(lr){8-10}\cmidrule(lr){11-13}\cmidrule(lr){14-16}
 & U & FK & \textbf{RG} & U & FK & \textbf{RG} & U & FK & \textbf{RG} & U & FK & \textbf{RG} & U & FK & \textbf{RG} \\
\midrule
\parbox[c][12.5pt][c]{\rgInstructionWidth}{\raggedright Place two mugs on plates} & \rgTableCell{21pt}{4.0} & \rgTableCell{21pt}{34.0} & \rgTableCell{21pt}{\textbf{56.0}} & \rgTableCell{21pt}{17.0} & \rgTableCell{21pt}{\textbf{38.0}} & \rgTableCell{21pt}{37.0} & \rgTableCell{21pt}{9.0} & \rgTableCell{21pt}{33.0} & \rgTableCell{21pt}{\textbf{42.0}} & \rgTableCell{21pt}{10.0} & \rgTableCell{21pt}{35.0} & \rgTableCell{21pt}{\textbf{45.0}} & \rgTableCell{21pt}{6.0} & \rgTableCell{21pt}{37.0} & \rgTableCell{21pt}{\textbf{55.0}} \\
\parbox[c][12.5pt][c]{\rgInstructionWidth}{\raggedright Place cream cheese in bowl} & \rgTableCell{21pt}{48.0} & \rgTableCell{21pt}{40.0} & \rgTableCell{21pt}{\textbf{63.0}} & \rgTableCell{21pt}{56.0} & \rgTableCell{21pt}{\textbf{62.0}} & \rgTableCell{21pt}{59.0} & \rgTableCell{21pt}{56.0} & \rgTableCell{21pt}{46.0} & \rgTableCell{21pt}{\textbf{65.0}} & \rgTableCell{21pt}{53.3} & \rgTableCell{21pt}{49.3} & \rgTableCell{21pt}{\textbf{62.3}} & \rgTableCell{21pt}{64.0} & \rgTableCell{21pt}{57.0} & \rgTableCell{21pt}{\textbf{79.0}} \\
\parbox[c][12.5pt][c]{\rgInstructionWidth}{\raggedright Move bowl from table center to plate} & \rgTableCell{21pt}{74.0} & \rgTableCell{21pt}{77.0} & \rgTableCell{21pt}{\textbf{98.0}} & \rgTableCell{21pt}{\textbf{95.0}} & \rgTableCell{21pt}{85.0} & \rgTableCell{21pt}{93.0} & \rgTableCell{21pt}{71.0} & \rgTableCell{21pt}{96.0} & \rgTableCell{21pt}{\textbf{97.0}} & \rgTableCell{21pt}{80.0} & \rgTableCell{21pt}{86.0} & \rgTableCell{21pt}{\textbf{96.0}} & \rgTableCell{21pt}{74.0} & \rgTableCell{21pt}{88.0} & \rgTableCell{21pt}{\textbf{96.0}} \\
\parbox[c][12.5pt][c]{\rgInstructionWidth}{\raggedright Place cream cheese and butter in basket} & \rgTableCell{21pt}{\textbf{87.0}} & \rgTableCell{21pt}{46.0} & \rgTableCell{21pt}{80.0} & \rgTableCell{21pt}{77.0} & \rgTableCell{21pt}{45.0} & \rgTableCell{21pt}{\textbf{80.0}} & \rgTableCell{21pt}{74.0} & \rgTableCell{21pt}{76.0} & \rgTableCell{21pt}{\textbf{86.0}} & \rgTableCell{21pt}{79.3} & \rgTableCell{21pt}{55.7} & \rgTableCell{21pt}{\textbf{82.0}} & \rgTableCell{21pt}{76.0} & \rgTableCell{21pt}{64.0} & \rgTableCell{21pt}{\textbf{83.0}} \\
\parbox[c][12.5pt][c]{\rgInstructionWidth}{\raggedright Place ketchup in basket} & \rgTableCell{21pt}{68.0} & \rgTableCell{21pt}{\textbf{94.0}} & \rgTableCell{21pt}{83.0} & \rgTableCell{21pt}{84.0} & \rgTableCell{21pt}{85.0} & \rgTableCell{21pt}{\textbf{89.0}} & \rgTableCell{21pt}{68.0} & \rgTableCell{21pt}{\textbf{90.0}} & \rgTableCell{21pt}{81.0} & \rgTableCell{21pt}{73.3} & \rgTableCell{21pt}{\textbf{89.7}} & \rgTableCell{21pt}{84.3} & \rgTableCell{21pt}{85.0} & \rgTableCell{21pt}{\textbf{98.0}} & \rgTableCell{21pt}{91.0} \\
\parbox[c][12.5pt][c]{\rgInstructionWidth}{\raggedright Turn on stove, place moka pot} & \rgTableCell{21pt}{59.0} & \rgTableCell{21pt}{89.0} & \rgTableCell{21pt}{\textbf{90.0}} & \rgTableCell{21pt}{60.0} & \rgTableCell{21pt}{\textbf{88.0}} & \rgTableCell{21pt}{82.0} & \rgTableCell{21pt}{79.0} & \rgTableCell{21pt}{\textbf{82.0}} & \rgTableCell{21pt}{78.0} & \rgTableCell{21pt}{66.0} & \rgTableCell{21pt}{\textbf{86.3}} & \rgTableCell{21pt}{83.3} & \rgTableCell{21pt}{60.0} & \rgTableCell{21pt}{78.0} & \rgTableCell{21pt}{\textbf{85.0}} \\
\parbox[c][12.5pt][c]{\rgInstructionWidth}{\raggedright Place orange juice in basket} & \rgTableCell{21pt}{92.0} & \rgTableCell{21pt}{\textbf{100.0}} & \rgTableCell{21pt}{\textbf{100.0}} & \rgTableCell{21pt}{\textbf{81.0}} & \rgTableCell{21pt}{75.0} & \rgTableCell{21pt}{79.0} & \rgTableCell{21pt}{80.0} & \rgTableCell{21pt}{\textbf{93.0}} & \rgTableCell{21pt}{91.0} & \rgTableCell{21pt}{84.3} & \rgTableCell{21pt}{89.3} & \rgTableCell{21pt}{\textbf{90.0}} & \rgTableCell{21pt}{88.0} & \rgTableCell{21pt}{98.0} & \rgTableCell{21pt}{\textbf{100.0}} \\
\parbox[c][12.5pt][c]{\rgInstructionWidth}{\raggedright Open top drawer, place bowl} & \rgTableCell{21pt}{1.0} & \rgTableCell{21pt}{\textbf{49.0}} & \rgTableCell{21pt}{48.0} & \rgTableCell{21pt}{41.0} & \rgTableCell{21pt}{57.0} & \rgTableCell{21pt}{\textbf{66.0}} & \rgTableCell{21pt}{45.0} & \rgTableCell{21pt}{45.0} & \rgTableCell{21pt}{\textbf{84.0}} & \rgTableCell{21pt}{29.0} & \rgTableCell{21pt}{50.3} & \rgTableCell{21pt}{\textbf{66.0}} & \rgTableCell{21pt}{8.0} & \rgTableCell{21pt}{\textbf{48.0}} & \rgTableCell{21pt}{\textbf{48.0}} \\
\parbox[c][12.5pt][c]{\rgInstructionWidth}{\raggedright Place bowl in bottom drawer, close it} & \rgTableCell{21pt}{51.0} & \rgTableCell{21pt}{\textbf{72.0}} & \rgTableCell{21pt}{69.0} & \rgTableCell{21pt}{64.0} & \rgTableCell{21pt}{66.0} & \rgTableCell{21pt}{\textbf{75.0}} & \rgTableCell{21pt}{63.0} & \rgTableCell{21pt}{\textbf{64.0}} & \rgTableCell{21pt}{57.0} & \rgTableCell{21pt}{59.3} & \rgTableCell{21pt}{\textbf{67.3}} & \rgTableCell{21pt}{67.0} & \rgTableCell{21pt}{53.0} & \rgTableCell{21pt}{78.0} & \rgTableCell{21pt}{\textbf{83.0}} \\
\parbox[c][12.5pt][c]{\rgInstructionWidth}{\raggedright Move bowl between plate and ramekin to plate} & \rgTableCell{21pt}{68.0} & \rgTableCell{21pt}{78.0} & \rgTableCell{21pt}{\textbf{94.0}} & \rgTableCell{21pt}{\textbf{82.0}} & \rgTableCell{21pt}{78.0} & \rgTableCell{21pt}{80.0} & \rgTableCell{21pt}{66.0} & \rgTableCell{21pt}{63.0} & \rgTableCell{21pt}{\textbf{80.0}} & \rgTableCell{21pt}{72.0} & \rgTableCell{21pt}{73.0} & \rgTableCell{21pt}{\textbf{84.7}} & \rgTableCell{21pt}{60.0} & \rgTableCell{21pt}{87.0} & \rgTableCell{21pt}{\textbf{93.0}} \\
\midrule
\parbox[c][12.5pt][c]{\rgInstructionWidth}{\raggedright Mean (all 10 tasks)} & \rgTableCell{21pt}{55.2} & \rgTableCell{21pt}{67.9} & \rgTableCell{21pt}{\textbf{78.1}} & \rgTableCell{21pt}{65.7} & \rgTableCell{21pt}{67.9} & \rgTableCell{21pt}{\textbf{74.0}} & \rgTableCell{21pt}{61.1} & \rgTableCell{21pt}{68.8} & \rgTableCell{21pt}{\textbf{76.1}} & \rgTableCell{21pt}{60.7} & \rgTableCell{21pt}{68.2} & \rgTableCell{21pt}{\textbf{76.1}} & \rgTableCell{21pt}{57.4} & \rgTableCell{21pt}{73.3} & \rgTableCell{21pt}{\textbf{81.3}} \\
\bottomrule
\end{tabular}
\endgroup
\end{table*}

\paragraph{5. Photometric sensitivity ($s_j^{\text{photo}}$).} To identify windows where the visuomotor mapping remains vulnerable to appearance variations, we probe each anchor $i$ with a fixed, deterministic camera transformation reused across refreshes, parameterized by brightness $\alpha$, contrast $\gamma$, and per-channel RGB gain $\mathbf g$:
\begin{equation}
I_i'=\operatorname{clip}_{[0,1]}\bigl(\alpha\,\mathbf g\odot[\gamma(I_i-\mu_i)+\mu_i]\bigr),
\end{equation}
where $\mu_i$ is the spatial image mean. Holding $c=(m,b)$ fixed, the policy generates action reconstructions $\hat{\mathbf a}_{i,c}$ from $I_i$ and $\hat{\mathbf a}'_{i,c}$ from $I_i'$, yielding:
\begin{equation}
\begin{gathered}
s_i^{\text{photo}}=\frac{1}{|\mathcal{C}|d_a|\mathcal M_i|}\!\sum_{c\in\mathcal{C},\,k\in\mathcal M_i}\!
\bigl\|\hat{\mathbf a}_{i,c}(k)-\hat{\mathbf a}'_{i,c}(k)\bigr\|_2^2,\\
\bar{s}_j^{\text{photo}}=\frac{1}{|W_j|}\sum_{i\in W_j}s_i^{\text{photo}},\qquad
s_j^{\text{photo}}=\log\frac{\bar{s}_j^{\text{photo}}+\epsilon}{\ell_j^{(r)}+\epsilon}.
\end{gathered}
\end{equation}
The raw window shift $\bar{s}_j^{\text{photo}}$ is normalized by current loss via the same log ratio as residual disagreement. Larger $s_j^{\text{photo}}$ indicates appearance sensitivity relative to denoising error. Unlike training data augmentation, $I_i'$ is a read-only probe that never enters gradient updates; it audits which original demonstration windows need additional exposure to stabilize visual representations.

\paragraph{Signal normalization and utility aggregation.} At each refresh, we recompute each signal's median and interquartile range ($\operatorname{IQR}$) across the task's $N_w$ windows, then standardize $s_j^k$ ($k\in\mathcal{S}$) with clipping bound $\kappa>0$:
\begin{equation}
z\bigl(s_j^k\bigr)=\operatorname{clip}\!\left(
\frac{s_j^k-\operatorname{median}_{l}s_l^k}
{\operatorname{IQR}_{l}\bigl(s_l^k\bigr)+\epsilon},-\kappa,\kappa\right).
\end{equation}
We combine the standardized signals with non-negative weights $\{w_k\}_{k\in\mathcal{S}}$ into an instantaneous utility $\tilde u_j^{(r)}=\sum_{k\in\mathcal{S}}w_k\,z(s_j^{k})$, and smooth across refreshes via an exponential moving average (EMA) with $\beta\in[0,1)$:
\begin{equation}
u_j^{(r)}=\beta u_j^{(r-1)}+(1-\beta)\tilde u_j^{(r)},
\end{equation}
initialized at $u_j^{(1)}=\tilde u_j^{(1)}$. All five signals are derived from offline demonstrations and policy states without rollouts.

\subsection{Rank-Based Sampling}
Directly converting raw utilities $u_j^{(r)}$ into sampling weights risks distribution collapse under outliers. Instead, let $r_j\in\{0,1,\dots,N_w-1\}$ denote the ascending rank of window $W_j$ under $u_j^{(r)}$ across the $N_w$ windows (ties averaged). Every valid anchor $i\in W_j$ inherits unnormalized weight $r_j+0.5$, which is normalized across all $N_a$ valid anchors to yield the adaptive distribution $q_i$. To prevent support starvation, we mix $q_i$ with a uniform coverage floor $\lambda\in(0,1)$:
\begin{equation}
p_i=\lambda\frac{1}{N_a}+(1-\lambda)q_i,\label{eq:sampling-probability}
\end{equation}
from which the data loader draws anchor indices $i\sim p$ with replacement.

\paragraph{Sampler properties and alternating loop.} Rank-based sampling with a uniform floor guarantees: (i)~\emph{nonzero coverage} ($p_i\geq \lambda/N_a$, bounding non-visitation after $N_{\text{draw}}$ draws by $\exp(-\lambda N_{\text{draw}}/N_a)$); (ii)~\emph{hyperparameter robustness and scale invariance} (because signals are standardized before aggregation and mapped to linear ranks mixed with $\lambda=0.5$, per-refresh window sampling probabilities are strictly bounded within $[0.5\times,1.5\times]$ of uniform without sensitivity to raw score spikes, allowing one fixed configuration across all tasks without per-task tuning); and (iii)~\emph{minimal state overhead} ($O(N_a+N_w)$ training memory and zero inference cost). After a uniform warmup recording $\ell_j^{(0)}$, \method alternates every $I_{\text{ref}}$ steps between read-only scoring and policy optimization on $p_i$ (step-by-step summary in \cref{sec:appendix-implementation}, \cref{tab:signal-definitions}).

%% file: sec/4_experiments.tex
\section{Experiments}
\label{sec:experiments}

\subsection{Benchmark, Implementation, and Baselines}
\begin{table}[!b]
\centering
\caption{Success (\%) by manipulation topology. Pert.\ averages lighting, noise, and texture; Clean uses unperturbed scenes. Families weight tasks equally. U: Uniform; FK: FAKTUAL; RG: \method; $\Delta$: gain in pp, computed before rounding.}
\label{tab:manipulation-topology}
\begingroup
\footnotesize
\setlength{\tabcolsep}{4.2pt}
\renewcommand{\arraystretch}{1.12}
\begin{tabular}{@{}p{87pt}lrr>{\columncolor{rgmethod}}rl@{}}
\toprule
Family (task IDs) & Scene & U & FK & \textbf{RG} & $\Delta$ vs.\ U / FK \\
\midrule
\multirow{2}{87pt}{\shortstack[l]{Single-object container\\(T1, T4, T6)}} & Pert. & 70.3 & 76.1 & \textbf{78.9} & +8.6 / +2.8 \\
 & Clean & 79.0 & 84.3 & \textbf{90.0} & +11.0 / +5.7 \\
\addlinespace[3pt]
\multirow{2}{87pt}{\shortstack[l]{Precision bowl transfer\\(T2, T9)}} & Pert. & 76.0 & 79.5 & \textbf{90.3} & +14.3 / +10.8 \\
 & Clean & 67.0 & 87.5 & \textbf{94.5} & +27.5 / +7.0 \\
\addlinespace[3pt]
\multirow{2}{87pt}{\shortstack[l]{Multi-object sequential\\(T0, T3)}} & Pert. & 44.7 & 45.3 & \textbf{63.5} & +18.8 / +18.2 \\
 & Clean & 41.0 & 50.5 & \textbf{69.0} & +28.0 / +18.5 \\
\addlinespace[3pt]
\multirow{2}{87pt}{\shortstack[l]{Articulated mechanism\\(T5, T7, T8)}} & Pert. & 51.4 & 68.0 & \textbf{72.1} & +20.7 / +4.1 \\
 & Clean & 40.3 & 68.0 & \textbf{72.0} & +31.7 / +4.0 \\
\bottomrule
\end{tabular}
\endgroup
\end{table}
\paragraph{Benchmark and protocol.} We evaluate ten LIBERO-Plus tasks (0, 1, 2, 3, 4, 5, 6, 7, 8, and 9), spanning container placement, bowl transfer, sequential manipulation, and articulated mechanisms. Each method receives 100 rollouts per task under lighting, noise, texture, and Clean conditions (4,000 total). Demonstrations already contain appearance variation. The policy uses no language input, isolating visuomotor allocation from language grounding; stratification criteria and task details appear in \cref{sec:appendix-implementation}.

\paragraph{Implementation and baselines.} All tasks share one fixed FP32 Diffusion Policy configuration (100,000 steps, batch size 8, $L_w=40$, $I_{\text{ref}}=10{,}000$, $\beta=0.5$, $\lambda=0.5$) without per-task tuning. We compare \method against \emph{Uniform} (equal anchor probability), \emph{FAKTUAL} \cite{sirigiri2026faktual} (offline trajectory-kernel curation), and controlled sampler ablations (\cref{tab:implementation-config,sec:appendix-controls}).

\input{sec/5_results}

%% file: sec/5_results.tex
\subsection{Main Results: Perturbed and Clean Success}
\label{sec:results}

\Cref{tab:refreshgrad-selected-10} reports success across the ten-task panel. Over 3,000 perturbed rollouts, \method reaches \textbf{76.1\%}, exceeding Uniform (60.7\%) by \textbf{15.4 pp} and FAKTUAL (68.2\%) by 7.9 pp. It improves perturbed success over Uniform on all ten tasks, with a median gain of +11.8 pp and up to +37.0 pp on Task 7 (29.0\% $\rightarrow$ 66.0\%).

\paragraph{Gains across manipulation topologies.} \Cref{tab:manipulation-topology} groups the panel by manipulation structure. Perturbed gains over Uniform rise from +8.6 pp for single-object container placement (Tasks 1, 4, 6) and +14.3 pp for precision bowl transfer (Tasks 2, 9) to \textbf{+18.8 pp} for multi-object sequential placement (Tasks 0, 3) and \textbf{+20.7 pp} for articulated mechanisms (Tasks 5, 7, 8; +31.7 pp Clean). Gains increase with temporal-stage complexity across these families.

\begin{table}[!t]
\centering
\caption{Ablation and sampler controls on Task 7 under lighting, noise, and texture perturbations. Entries report success rates.}
\label{tab:task7-ablation}
\footnotesize
\setlength{\tabcolsep}{3.2pt}
\begin{tabular}{@{}lrrrr@{}}
\toprule
Sampler Variant & Lighting & Noise & Texture & Overall \\
\midrule
\rowcolor{rgfull}
Uniform & 1.0\% & 41.0\% & 45.0\% & 29.0\% \\
\rowcolor{rgmethod}
\textbf{\method (complete)} & 48.0\% & \textbf{66.0\%} & \textbf{84.0\%} & \textbf{66.0\%} \\
\midrule
No gradient agreement ($s_j^{\text{grad}}$) & 17.0\% & 52.0\% & 47.0\% & 38.7\% \\
No photometric sensitivity ($s_j^{\text{photo}}$) & 36.0\% & 49.0\% & 61.0\% & 48.7\% \\
No temporal consistency ($s_j^{\text{temp}}$) & 47.0\% & 47.0\% & 53.0\% & 49.0\% \\
No residual disagreement ($s_j^{\text{res}}$) & 50.0\% & 47.0\% & 52.0\% & 49.7\% \\
Loss-progress only (EMA 0.0) & 47.0\% & 58.0\% & 62.0\% & 55.7\% \\
Loss-progress only (EMA 0.5) & 50.0\% & 61.0\% & 57.0\% & 56.0\% \\
\midrule
Loss-magnitude priority & 43.0\% & 47.0\% & 51.0\% & 47.0\% \\
\bottomrule
\end{tabular}
\end{table}

\begin{table}[!b]
\centering
\caption{Success rates for progress-only sampling (EMA 0.5) versus full \method across all ten tasks (300 perturbed evaluation episodes per task; 3,000 total).}
\label{tab:progress-ten}
\footnotesize
\begin{tabular}{@{}l >{\columncolor{rgfull}}c c >{\columncolor{rgmethod}}c@{}}
\toprule
Task & Uniform & Progress-only & \textbf{\method} \\
\midrule
0 & 10.0\% & 44.7\% & \textbf{45.0\%} \\
1 & 53.3\% & 57.7\% & \textbf{62.3\%} \\
2 & 80.0\% & 95.3\% & \textbf{96.0\%} \\
3 & 79.3\% & 68.3\% & \textbf{82.0\%} \\
4 & 73.3\% & \textbf{91.0\%} & 84.3\% \\
5 & 66.0\% & \textbf{83.7\%} & 83.3\% \\
6 & 84.3\% & 76.7\% & \textbf{90.0\%} \\
7 & 29.0\% & 56.0\% & \textbf{66.0\%} \\
8 & 59.3\% & 58.7\% & \textbf{67.0\%} \\
9 & 72.0\% & 65.3\% & \textbf{84.7\%} \\
\midrule
All & 60.7\% & 69.7\% & \textbf{76.1\%} \\
\bottomrule
\end{tabular}
\end{table}

\begin{figure*}[!t]
\centering
\includegraphics[width=\textwidth]{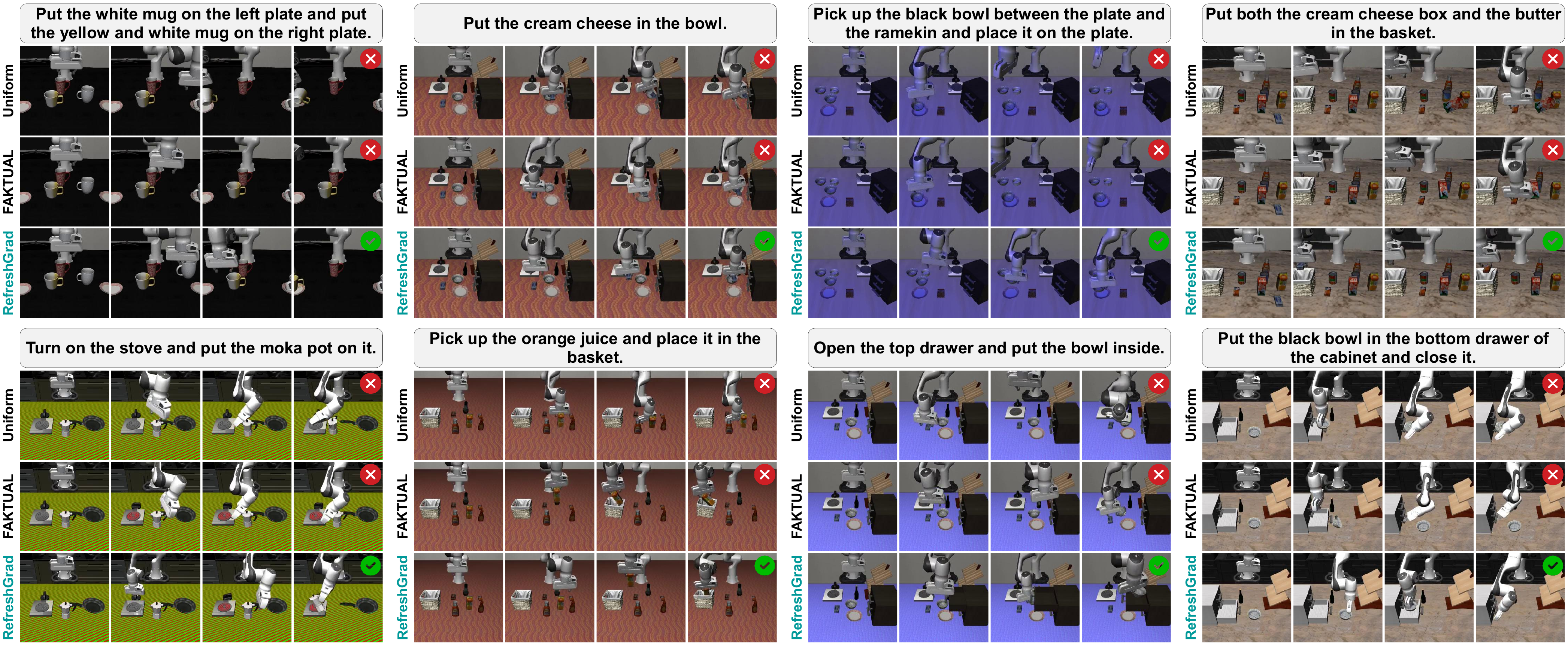}
\caption{\textbf{Qualitative comparison across eight representative tasks under visual perturbations.} Each panel compares matched rollouts of Uniform, FAKTUAL, and \method (frames progress left to right). Only \method completes the tasks across these perturbed episodes.}
\label{fig:qualitative-comparison}
\end{figure*}

On multi-object sequential tasks, two grasp--place cycles occur within one episode. FAKTUAL reaches 45.3\%, close to Uniform's 44.7\%, whereas \method reaches \textbf{63.5\%}: \textbf{+18.2 pp} over FAKTUAL under perturbations and +18.5 pp in Clean scenes. This gap is consistent with dynamic window allocation isolating bottleneck transitions within episodes, which trajectory curation cannot target separately.

\paragraph{Perturbation-wise and clean performance.} \method leads both baselines not only under lighting (+22.9 pp over Uniform), but also under non-photometric texture shifts (+15.0 pp) and sensor noise (+8.3 pp) (\cref{tab:refreshgrad-selected-10}). Moreover, in unperturbed Clean scenes, success rises from 57.4\% (Uniform) and 73.3\% (FAKTUAL) to \textbf{81.3\%} (+23.9 / +8.0 pp), improving over Uniform on all ten tasks and confirming that gains do not rely on matching a test-time corruption.

\subsection{Mechanism Analysis and Ablations}
\paragraph{Allocation patterns and value dynamics.} Cumulative window exposure relative to uniform sampling spans $0.53\times$--$1.58\times$ across tasks (up to $2.40\times$ for bottleneck windows), versus $0.69\times$--$1.33\times$ at episode level; 25.6\% of anchors lie outside $[0.8\times,1.2\times]$. These are allocation outcomes: the five cues induce selective within-episode reweighting finer than trajectory-level allocation. Independently, Uniform-training logs show window-signal heterogeneity and rank drift without adaptive sampling (\cref{fig:signal-decay}(b)); diagnostic details are in \cref{sec:appendix-diagnostics}.

\paragraph{Contribution of individual signals and sampler controls.} We evaluate signal contributions at two levels: a ten-task ablation of all four auxiliary cues (\cref{tab:progress-ten}) and a leave-one-signal-out breakdown on the articulated Task 7 benchmark (\cref{tab:task7-ablation}). In \cref{tab:task7-ablation}, removing gradient agreement reduces Task 7 success from 66.0\% to 38.7\%; removing photometric sensitivity, temporal consistency, or residual disagreement yields 48.7\%, 49.0\%, or 49.7\%, respectively (all still +9.7 to +20.7 pp above Uniform's 29.0\%). Each cue also targets distinct failure modes: photometric removal lowers lighting success from 48.0\% to 36.0\%, while temporal removal lowers noise success from 66.0\% to 47.0\%. Under the identical rank-based sampler, warmup, and uniform floor, prioritizing raw current loss (\emph{Loss-magnitude priority}) reaches only 47.0\%, whereas tracking loss reduction (\emph{Loss-progress only}) reaches 55.7\%--56.0\% and full \method reaches \textbf{66.0\%}, confirming that gains stem from multi-cue policy feedback rather than raw loss magnitude (\cref{sec:appendix-signal-analysis}).

\paragraph{Progress-only versus full utility across ten tasks.} Across all ten tasks (\cref{tab:progress-ten}, 3,000 rollouts), learning progress alone lifts perturbed success from 60.7\% to 69.7\% (+9.1 pp), while adding the four consistency and sensitivity signals yields a further \textbf{+6.3 pp} gain to \textbf{76.1\%}, winning on 8 of 10 tasks (+19.3 pp on Task 9, +13.7 pp on Task 3, and +13.3 pp on Task 6). Progress-only already saturates on simpler high-baseline tasks where demonstrations are largely conflict-free (Tasks 4 and 5), whereas the four auxiliary signals are decisive on multi-stage and contact-sensitive tasks.

\paragraph{Refresh overhead and modularity.} Because all scoring signals are extracted directly from the current policy without external proxy networks, \method decouples data allocation from the policy architecture and training objective. With four deterministic scoring conditions per anchor and gradient agreement restricted to the final projection head, refresh scoring costs 15.7\% of wall time in the controlled Task 7 run and 19\% median across tasks, with \textbf{zero inference overhead} (\cref{sec:appendix-overhead}).

\paragraph{Qualitative comparison.} \Cref{fig:qualitative-comparison} compares matched rollouts across eight tasks under lighting and texture perturbations. Uniform and FAKTUAL stall or miss transitions, whereas \method completes all eight tasks (\cref{tab:refreshgrad-selected-10}).

%% file: sec/appendix.tex
\clearpage
\renewcommand{\thepage}{S\arabic{page}}
\setcounter{page}{1}
\maketitlesupplementary
\appendix

\section{Method Summary and Implementation Details}
\label{sec:appendix-implementation}

\paragraph{Alternating training loop and component summary.}
Training starts with a uniform warmup pass over all valid anchors, during which baseline losses $\ell_j^{(0)}$ are recorded halfway through warmup. Thereafter, every $I_{\text{ref}}$ optimizer steps, \method executes a four-stage refresh transition:
\begin{enumerate}[leftmargin=*,itemsep=1pt]
\item \textbf{Score:} Evaluate the five policy-native signals $\{s_j^{k}\}_{k\in\mathcal{S}}$ across eligible windows using the current policy $\pi_\theta$ and deterministic scoring probes $\mathcal{C}$.
\item \textbf{Aggregate:} Apply robust standardization $z(\cdot)$ and update the smoothed window utilities $u_j^{(r)}$.
\item \textbf{Allocate:} Compute rank-based weights $q_i$ and update the mixture distribution $p_i$ in \cref{eq:sampling-probability}.
\item \textbf{Train:} Resume standard policy optimization by drawing anchors from $p_i$ and minimizing the unchanged per-sample loss $\mathcal{L}(\theta;i)$.
\end{enumerate}
Because the scoring pass is decoupled from gradient updates, the sampler state, historical utilities, and anchor probabilities are serialized alongside model and optimizer checkpoints. \Cref{tab:signal-definitions} summarizes the functional role of each signal and sampler component.

\begin{table}[!t]
\centering
\caption{Summary of policy-native signals and sampler components in \method.}
\label{tab:signal-definitions}
\footnotesize
\setlength{\tabcolsep}{3pt}
\begin{tabular}{@{}lp{4.6cm}@{}}
\toprule
Component & Role in window allocation \\
\midrule
Learning progress ($s_j^{\text{prog}}$) & Prioritizes windows where denoising error is actively decreasing between refreshes. \\
Gradient agreement ($s_j^{\text{grad}}$) & Verifies that window gradients align with visually similar windows across episodes. \\
Temporal consistency ($s_j^{\text{temp}}$) & Rewards coherent action-chunk predictions across adjacent overlapping anchors. \\
Residual disagreement ($s_j^{\text{res}}$) & Captures inter-noise-bank prediction variance normalized by current loss scale. \\
Photometric sensitivity ($s_j^{\text{photo}}$) & Measures photometric action-prediction shifts as a log ratio to current window loss. \\
Rank-based sampling ($q_i$) & Converts utility ranks across the task's windows into adaptive anchor weights without sensitivity to raw score scale. \\
Uniform mixture ($p_i$) & Guarantees nonzero sampling support ($\ge \lambda/N_a$) and global coverage for all valid anchors. \\
\bottomrule
\end{tabular}
\end{table}

\paragraph{Stratified ten-task panel details.}
To rigorously test temporal-window allocation across diverse horizon lengths and contact structures, our ten-task LIBERO-Plus panel (Tasks 0, 1, 2, 3, 4, 5, 6, 7, 8, and 9) is guided by three criteria:
(i)~\emph{Visual task identifiability}: because our Diffusion Policy instantiation receives no language conditioning in order to isolate visuomotor learning from language-grounding confounds, each selected task has a scene layout that visually defines the manipulation objective without cross-task instruction ambiguity;
(ii)~\emph{Stratified manipulation-structure topology}: the panel spans four distinct manipulation structures---single-object container placement (Tasks 1, 4, 6), precision workspace bowl transfer (Tasks 2, 9), multi-object sequential placement requiring two consecutive grasp--place cycles within one episode (Tasks 0, 3), and multi-stage articulated mechanism manipulation chaining stove actuation or drawer opening/closing with object placement (Tasks 5, 7, 8); and
(iii)~\emph{Baseline difficulty coverage}: nominal uniform-training perturbed performance spans from 10.0\% and 29.0\% on Tasks 0 and 7 to 84.3\% on Task 6.
Note that the photometric probe used during scoring (\cref{sec:method}) applies a fixed brightness, contrast, and RGB-gain shift solely for utility estimation and is independent of the evaluation corruptions (where image noise also includes blur-type sensor degradations). Because the evaluated policy receives no language input, LIBERO-Plus unperturbed variants (which differ from base tasks only in instruction phrasing) directly measure clean visuomotor competence.

\paragraph{Complete hyperparameter configuration.}
\Cref{tab:implementation-config} lists the complete hyperparameter instantiation of \method, which is held fixed across all evaluated tasks without per-task tuning. Each episode is partitioned into windows of length $L_w=40$ anchors, and refreshes occur every $I_{\text{ref}}=10{,}000$ optimizer steps. The deterministic scoring condition set $\mathcal{C}$ pairs two diffusion timesteps $\mathcal{T}_s=\{T/4,3T/4\}$ with $B=2$ fixed frame-aligned noise banks ($|\mathcal{C}|=4$). Cross-episode gradient agreement ($s_j^{\text{grad}}$) is computed over the final $1\times 1$ action projection layer using $M=16$ visually nearest neighbor windows and a coordinate-wise trim ratio $\rho=0.2$. Photometric sensitivity ($s_j^{\text{photo}}$) samples deterministic per-anchor brightness and contrast factors $\alpha,\gamma\in[0.6,1.4]$ and RGB channel gains $\mathbf g\in[0.85,1.15]^3$. Robust standardization uses clip threshold $\kappa=3$ and numerical constant $\epsilon=10^{-12}$; standardized signals are combined with weights $(w_{\text{prog}},w_{\text{grad}},w_{\text{temp}},w_{\text{res}},w_{\text{photo}})=(1.0,0.5,0.25,0.25,0.25)$, smoothed with EMA factor $\beta=0.5$, and mixed with the uniform distribution using $\lambda=0.5$.

\begin{table}[!t]
\centering
\caption{\method hyperparameter and training configuration shared across all tasks.}
\label{tab:implementation-config}
\footnotesize
\setlength{\tabcolsep}{4pt}
\begin{tabular}{@{}ll@{}}
\toprule
Parameter (Symbol) & Value / Configuration \\
\midrule
Policy backbone ($\pi_\theta$) & FP32 Diffusion Policy \\
Training budget \& batch size & 100,000 steps, batch size 8 \\
Temporal window length ($L_w$) & 40 consecutive anchors \\
Refresh interval ($I_{\text{ref}}$) & 10,000 optimizer steps \\
Scoring timesteps ($\mathcal{T}_s$) & $\{T/4,\, 3T/4\}$ ($|\mathcal{T}_s|=2$) \\
Fixed noise banks ($B$) & 2 banks ($|\mathcal{C}|=4$ conditions) \\
Gradient neighborhood ($M, \rho$) & $M=16$ neighbors, $\rho=0.2$ trim \\
Photometric probe ($\alpha, \gamma, \mathbf g$) & $\alpha,\gamma\in[0.6,1.4]$, $\mathbf g\in[0.85,1.15]^3$ \\
Standardization clip ($\kappa, \epsilon$) & $\kappa=3$, $\epsilon=10^{-12}$ \\
Signal weights ($\{w_k\}_{k\in\mathcal{S}}$) & $(1.0, 0.5, 0.25, 0.25, 0.25)$ \\
Utility EMA factor ($\beta$) & $0.5$ \\
Uniform mixture floor ($\lambda$) & $0.5$ \\
Data loader \& cache & 8 workers, prefetch 4, lossless RGB \\
\bottomrule
\end{tabular}
\end{table}

\subsection{Controlled Sampler Configurations}
\label{sec:appendix-controls}

\paragraph{Implementation and baselines.} All tasks share a single fixed hyperparameter configuration without per-task tuning (full table in \cref{sec:appendix-implementation}, \cref{tab:implementation-config}): episodes are partitioned into windows of $L_w=40$ anchors and refreshed every $I_{\text{ref}}=10{,}000$ steps over a 100,000-step FP32 Diffusion Policy run (batch size 8). Scoring uses $|\mathcal{C}|=4$ deterministic conditions ($\mathcal{T}_s=\{T/4,3T/4\}$, $B=2$ noise banks), $M=16$ cross-episode neighbors with trim $\rho=0.2$, photometric bounds $\alpha,\gamma\in[0.6,1.4]$ and $\mathbf g\in[0.85,1.15]^3$, standardization clip $\kappa=3$, signal weights $(1.0,0.5,0.25,0.25,0.25)$, EMA $\beta=0.5$, and uniform floor $\lambda=0.5$.
We compare \method against: (i)~\emph{Uniform}, which trains the identical architecture, steps, and seed under uniform anchor sampling; (ii)~\emph{FAKTUAL} \cite{sirigiri2026faktual}, an offline trajectory-kernel entropy curation baseline evaluated under the exact same protocol; and (iii)~three controlled sampler ablations holding the refresh schedule, rank-based sampling, and uniform floor fixed while varying only the utility definition: \emph{Progress-only} ($s_j^{\text{prog}}$ alone, $\beta\in\{0.0,0.5\}$), \emph{Loss-magnitude priority} (ranking by current window loss $\ell_j^{(r)}$), and \emph{Leave-one-signal-out} ablations of $s_j^{\text{grad}}, s_j^{\text{temp}}, s_j^{\text{res}}, s_j^{\text{photo}}$.

\section{Additional Diagnostics and Ablation Analysis}
\label{sec:appendix-results}

\subsection{Training-Trace Diagnostics}
\label{sec:appendix-diagnostics}

\begin{figure*}[t]
\centering
\includegraphics[width=\textwidth]{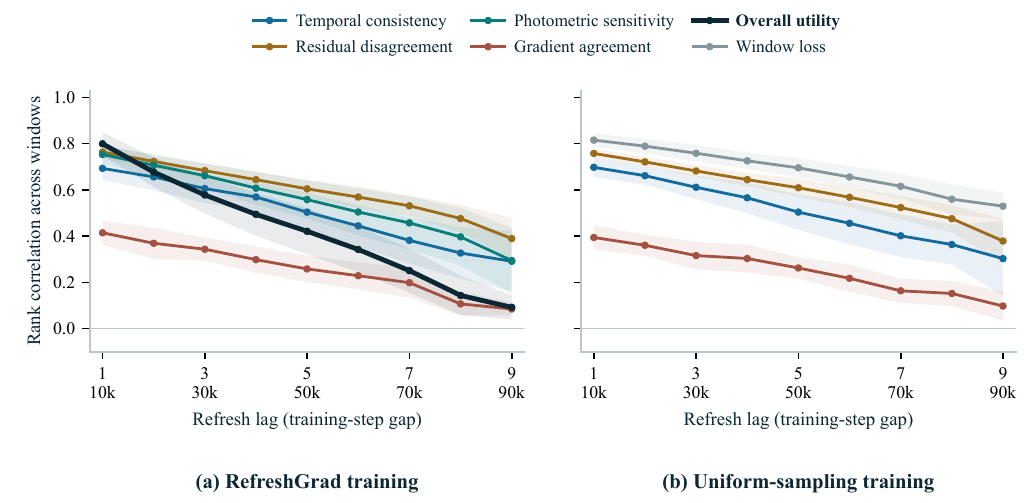}
\caption{Spearman rank correlation across windows between refreshes as a function of refresh lag, averaged equally over tasks (shaded: $\pm1$ SD across tasks; lags 1--8: ten tasks, lag 9: eight tasks). \textbf{(a)} For this drift diagnostic, Overall utility fixes normalization statistics at the first refresh while retaining EMA. Main and ablation runs (\crefrange{tab:refreshgrad-selected-10}{tab:progress-ten}) instead recompute medians/IQRs at every refresh; fixing them here provides a common scale for measuring cross-refresh rank drift. \textbf{(b)} Sampling is uniform over valid anchors, and photometric sensitivity is omitted. Residual disagreement and photometric sensitivity use log ratios to current loss before standardization. Rank persistence declines with lag under both training schemes.}
\label{fig:signal-decay}
\end{figure*}

\paragraph{Allocation outcomes and independent signal diagnostics.} The refresh logs separate sampler behavior from evidence about window-signal variation:
(i)~\emph{Five-cue allocation differentiates windows within episodes}: across ten tasks, cumulative window exposure relative to uniform sampling spans a mean range of $0.53\times$ to $1.58\times$ (up to $2.40\times$ for bottleneck windows), versus $0.69\times$ to $1.33\times$ at episode level. In total, $25.6\%$ of anchors fall outside $[0.8\times,1.2\times]$ ($13.5\%$ below $0.8\times$, $12.1\%$ above $1.2\times$). This is an outcome of five-signal reweighting: window allocation distinguishes segments within a trajectory more finely than episode-level allocation. Independent evidence of non-uniform, evolving window signals comes from Uniform-training logs in \cref{fig:signal-decay}(b), where adaptive sampling is absent.
(ii)~\emph{Policy-native signals capture variation beyond raw loss}: across 98 refresh passes, window loss $\ell_j^{(r)}$ correlates negatively with gradient agreement $s_j^{\text{grad}}$ (mean $r=-0.31$, reaching $-0.58$) and temporal consistency $s_j^{\text{temp}}$ (mean $r=-0.40$, reaching $-0.61$). Raw inter-bank disagreement correlates strongly with loss ($r=0.84$), whereas its loss-normalized log ratio $s_j^{\text{res}}$ has $r=0.13$. Raw photometric shift $\bar{s}_j^{\text{photo}}$ has $r=0.08$ with loss; after log-ratio normalization, $s_j^{\text{photo}}$ has $r=-0.20$. These associations support complementary sensitivity cues without equating either score with loss magnitude.
(iii)~\emph{Window utilities drift over training}: with normalization statistics fixed at the first refresh for this diagnostic (\cref{sec:appendix-value-dynamics}), utilities have task-equal mean adjacent-refresh Spearman correlation $\rho=0.80$ (88 pairs; median $0.83$, range $0.53$--$0.96$), falling to $0.10$ from first to last refresh. Together with the Uniform-training traces, this motivates online reassessment.

\subsection{Window-Value Dynamics During Training}
\label{sec:appendix-value-dynamics}

Scoring uses fixed denoising noise banks and timesteps, with fixed photometric perturbations when that signal is included.
Under \method, adjacent-refresh rank correlations are 0.69 for temporal consistency, 0.76 for residual disagreement, 0.75 for photometric sensitivity, and 0.80 for Overall utility, arguing against the changes being pure independent scoring noise.
These correlations decline with lag to first--last task means of 0.29, 0.41, 0.34, and 0.10, respectively; every displayed measure has a negative slope on all ten tasks, with window-trajectory bootstrap 95\% intervals below zero (1,000 resamples).
Under uniform sampling, residual disagreement similarly declines from 0.76 to 0.41 and window loss from 0.82 to 0.52, indicating that the dynamics accompany training beyond adaptive sampling.
The historical uniform runs use an earlier scoring formulation, so comparisons between training schemes are descriptive; bootstrap intervals are conditional on windows rather than independent training repeats.

\subsection{Signal Contributions and Control Interpretations}
\label{sec:appendix-signal-analysis}

\paragraph{Contribution of individual signals and sampler controls.} In \cref{tab:task7-ablation}, removing any of the four self-consistency and sensitivity signals degrades Task 7 performance from 66.0\%. Cross-episode gradient agreement ($s_j^{\text{grad}}$) is the single most impactful cue (-27.3 pp drop to 38.7\% when ablated). Removing photometric sensitivity ($s_j^{\text{photo}}$), temporal consistency ($s_j^{\text{temp}}$), or residual disagreement ($s_j^{\text{res}}$) reduces success to 48.7\%, 49.0\%, and 49.7\%, with each cue protecting against targeted failure modes (dropping $s_j^{\text{photo}}$ lowers lighting success from 48.0\% to 36.0\%; dropping $s_j^{\text{temp}}$ lowers noise success from 66.0\% to 47.0\%). Learning progress ($s_j^{\text{prog}}$) alone reaches 55.7\% (EMA 0.0) and 56.0\% (EMA 0.5) on Task 7; across all ten tasks (\cref{tab:progress-ten}), progress-only sampling lifts perturbed success from 60.7\% to 69.7\% (+9.1 pp), while adding the four self-consistency signals yields a further \textbf{+6.3 pp} gain to \textbf{76.1\%} (winning on 8 of 10 tasks, including +19.3 pp on Task 9 and +13.7 pp on Task 3).
\paragraph{Loss-priority and progress-only controls.} Holding the warmup pass, refresh schedule, rank-based weight set $\{0.5,1.5,\dots,N_w-0.5\}$, uniform floor $\lambda=0.5$, and sampling with replacement identical to \method, prioritizing windows by current loss magnitude (\emph{Loss-magnitude priority}) reaches only 47.0\% on Task 7 (-19.0 pp vs.\ \method), confirming that raw loss over-exposes the policy to noisy or conflicting transitions. Replacing raw loss with loss reduction (\emph{Loss-progress only}) raises success to 55.7\%--56.0\%, and incorporating the four consistency and sensitivity signals further elevates performance to 66.0\% (+10.0 pp), isolating the benefit of multi-cue policy-native valuation.

\section{Computational Overhead Analysis}
\label{sec:appendix-overhead}

\begin{table}[!t]
\centering
\caption{Measured refresh-scoring overhead across the ten LIBERO-Plus tasks.}
\label{tab:runtime-analysis}
\footnotesize
\setlength{\tabcolsep}{4pt}
\begin{tabular}{@{}lr@{}}
\toprule
Quantity & Measurement \\
\midrule
Refresh interval & 10,000 optimizer steps \\
Scoring passes per 100k-step run & 9--10 \\
Median eligible windows per task & 1,548 (range 960--3,225) \\
Median cumulative scoring time per task & 3,658 s (61.0 min) \\
Scoring share of training wall time & 10--30\% (median 19\%) \\
Controlled run, Task 7 (single RTX 6000D) & 1,190 s of 2.1 h (15.7\%) \\
Inference-time overhead & \textbf{0\% (unchanged policy)} \\
\bottomrule
\end{tabular}
\end{table}

At each 10,000-step refresh, \method scores eligible windows without policy updates and refreshes anchor probabilities. \Cref{tab:runtime-analysis} summarizes costs across ten tasks. Four deterministic conditions per anchor and gradients restricted to the final $1\times1$ projection head limit overhead. On one NVIDIA RTX 6000D, a complete 100k-step Task 7 run takes 2.1 hours, including 1,190 seconds of scoring (15.7\%; $\sim$110 seconds per refresh). Across tasks, scoring accounts for a median of 19\% (range 10--30\%) of training wall time, with zero inference overhead.